\pdfoutput=1
\documentclass[twoside,a4paper]{scrartcl}

\usepackage{graphicx}

\usepackage{hyperref}
   \hypersetup{
      pdftitle={Validation of nonlinear PCA},
      pdfsubject={Auto-associative neural networks},
      pdfauthor={Matthias Scholz},
      pdfkeywords={NLPCA, nonlinear PCA, neural networks, model selection, validation},
      plainpages=false,
      pdfstartview=Fit,  
      pdfstartpage=1,
      breaklinks=true,
      colorlinks=true, 
      pdfhighlight=/N,
      linkcolor=black,    
      citecolor=black,
      urlcolor=blue,
      citebordercolor=1 1 1,
      filebordercolor=1 1 1,
      linkbordercolor=1 1 1,
      menubordercolor=1 1 1,
      pagebordercolor=1 1 1,
      urlbordercolor=1 1 1,
      pdfborder=1 1 1
      }

\usepackage[numbers]{natbib} 
    \setcapwidth[c]{0.9\linewidth}
    \setcapindent{0em}
    \addtokomafont{caption}{\small}

\newenvironment{memo}
    {\begin{list}{}{\footnotesize%
    \setlength{\leftmargin}{0.0cm}%
    \setlength{\rightmargin}{0.0cm}}%
    \item[]\ignorespaces}%
    {\unskip\end{list}}

\begin{document}

\title{Validation of nonlinear PCA}
\date{}
\author{Matthias Scholz}

\publishers{\small 
  Edmund Mach Foundation -- Research and Innovation Center\\ 
  Via Edmund Mach 1, 38010~San~Michele all'Adige (TN), Italy \\   
  matthias.scholz@fmach.it
}

\maketitle

\vspace{-8.5cm} 
\begin{memo}
Matthias Scholz. Validation of nonlinear PCA. (pre-print version)
\hfill \href{http://www.NLPCA.org}{\texttt{www.NLPCA.org}}\linebreak[4] 
The final publication is available at www.springerlink.com\\
To appear in \emph{Neural Processing Letters,} 2012. \\
Doi: \href{http://dx.doi.org/10.1007/s11063-012-9220-6}{10.1007/s11063-012-9220-6}
\end{memo}
\vspace{5.6cm}

\begin{abstract}
\noindent
Linear principal component analysis (PCA) can be extended to a nonlinear PCA by using artificial neural networks. But the benefit of curved components requires a careful control of the model complexity. Moreover, standard techniques for model selection, including cross-validation and more generally the use of an independent test set, fail when applied to nonlinear PCA because of its inherent unsupervised characteristics.
\\
This paper presents a new approach for validating the complexity of nonlinear PCA models by using the error in missing data estimation as a criterion for model selection. It is motivated by the idea that only the model of optimal complexity is able to predict missing values with the highest accuracy. While standard test set validation usually favours over-fitted nonlinear PCA models, the proposed model validation approach correctly selects the optimal model complexity.
\end{abstract}

\section{Introduction}
\label{intro}
Nonlinear principal component analysis\,\cite{Kra91,DemCot93,Hec95} is a nonlinear generalization of standard principal component analysis (PCA).
While PCA is restricted to linear components, nonlinear PCA generalizes the principal components from straight lines to curves and hence describes the inherent structure of the data by curved subspaces.
Detecting and describing nonlinear structures is especially important for analysing time series. Nonlinear PCA is therefore frequently used to investigate the dynamics of different natural processes\,\cite{Hsi06,Her07,SchFra08}. But validating the model complexity of nonlinear PCA is a difficult task\,\cite{Chr05}.
Over-fitting can be caused by the often limited number of available samples; moreover, in nonlinear PCA over-fitting can also occur by the intrinsic geometry of the data, as shown in Fig.\,\ref{fig:GaussData}, which cannot be solved by increasing the number of samples.
A good control of the complexity of the nonlinear PCA model is required.
We have to find the optimal flexibility of the curved components.
A component with too little flexibility, an almost linear component, 
cannot follow the complex curved trajectory of real data. 
By contrast, a too flexible component fits non-relevant noise of the data (over-fitting) and hence gives a poor approximation of the original process, as illustrated in Fig.\,\ref{fig:crossval_illu}A.
The objective is to find a model whose complexity is neither too small nor too 
large.
\\
Even though the term nonlinear PCA (NLPCA) is often referred to 
the auto-associative neural network approach, there are many other methods which 
visualise data and extract components in a nonlinear manner\,\cite{GorKegWun07}. 
{\em Locally linear embedding} (LLE)\,\cite{RowSau00,SauRow03}
and {\em Isomap}\,\cite{TenSilLan00} visualise high dimensional
data by projecting (embedding) them into a two or three-dimensional space. 
{\em Principal curves} \cite{HasStu89} and
{\em self organising maps} (SOM)\,\cite{Koh01} describe data by nonlinear curves and nonlinear planes up to two dimensions.
{\em Kernel PCA}\,\cite{SchSmoMue98} as a kernel approach can be used to visualise data and for noise reduction\,\cite{MikSchSmo99}. 
In\,\cite{GirIov05} linear subspaces of PCA are replaced by manifolds and in\,\cite{DemHer97} a neural network approach is used for nonlinear mapping.
This work is focused on the auto-associative neural network approach to nonlinear PCA and its model validation problem.
\\
For supervised methods, a standard validation technique is cross-validation. 
But even though the neural network architecture used is supervised, the nonlinear PCA itself is an unsupervised method that requires validating techniques different from those used for supervised methods.  
A common approach for validating unsupervised methods is to validate the robustness of the components under moderate modifications of the original data set, e.g., 
by using resampling \emph{bootstrap}\,\cite{EfrTib94}
or by corrupting the data with a small amount of Gaussian noise\,\cite{HarMeiMue04}.
In both techniques, the motivation is that reliable components should be robust and 
stable against small random modification of the data.
In principle, these techniques could be adapted to nonlinear methods.
But there would be the difficulty of measuring the robustness of nonlinear components.
Robustness of linear components is measured by comparing their directions under slightly different conditions (resampled data sets or different noise-injections). 
But since comparing the curvature of nonlinear components is no trivial task, 
nonlinear methods require other techniques for model validation. 
\\
In a similar neural network based nonlinear PCA model, termed nonlinear factor analysis (NFA)\,\cite{HonVal05}, 
a Bayesian framework is used in which the weights and 
inputs are described by posterior probability distributions which leads to a good regularisation.
While in such Bayesian learning the inputs (components) are explicitly modelled by Gaussian 
distributions, the maximum likelihood approach in this work attempts to find a single set of values for the network 
weights and inputs. A weight-decay regulariser is used to control the model complexity. 
There are several attempts to the model selection in the auto-associative nonlinear PCA. Some are based on a criterion of how good the local neighbour relation is preserved by the nonlinear PCA transformation\,\cite{ChaGir99}. In\,\cite{Hsi07}, a nearest neighbour inconsistency term that penalises complex models is added to the error function, but standard test set validation is used for model pre-selection.
In\,\cite{LuPan11} an alternative network architecture is proposed to solve the problems of over-fitting and non-uniqueness of nonlinear PCA solutions.
Here we consider a natural approach that validates the model by its own ability to estimate missing data. Such missing data validation is used, e.g., for validating linear PCA models\,\cite{IliRai10}, and for comparing probabilistic nonlinear PCA models based on Gaussian processes\,\cite{Law05}. Here, the missing data validation approach is adapted to validate the auto-associative neural network based nonlinear PCA.

\section{The test set validation problem}
\label{sec:1}
\begin{figure}[t]
  \centerline{\includegraphics[width=1.0\linewidth]{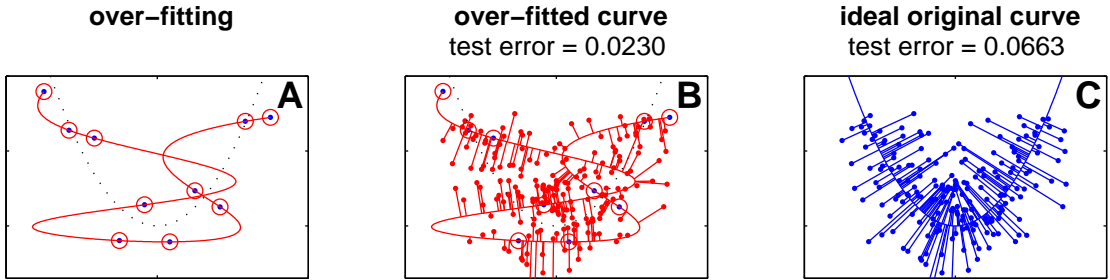}}
   \caption[abc]{
    The problem of mis-validation of nonlinear PCA by using a test data set. 
    Ten training samples were generated from a 
    quadratic function (dotted line) plus noise. 
    ({\sffamily\bfseries A})~A nonlinear PCA model of too high complexity 
    leads to an overfitted component (solid curve).
    But validating this over-fitted model with an independent test data 
    set~({\sffamily\bfseries B}) gives a better (smaller) test error than 
    using the original model from which the 
    data were generated~({\sffamily\bfseries C}).
    }\label{fig:crossval_illu}
 \end{figure}

To validate supervised methods, the standard approach is to use an independent test set for controlling the complexity of the model.
This can be done either by using a new data set, or when the number of samples is limited, by 
performing cross-validation by repeatedly splitting the original data into a training and test set.
The idea is that only the model, which best represents the 
underlying process, can provide optimal results on new, 
for the model previously unknown, data.
But test set validation only works well when there exist a clear target value (e.g., class labels) as in supervised methods, it fails on unsupervised methods.
In the same way that a test data set cannot be used to validate the optimal number of components in standard linear PCA, test data also cannot be used to validate the curvature of components in nonlinear PCA\,\cite{Chr05}.
Even though nonlinear PCA can be performed by using a supervised neural network architecture, it is still an unsupervised method and hence should not be validated by using cross-validation.
With increasing complexity, nonlinear PCA is able to provide a curved component with better data space coverage. Thus, also test data can be projected onto the (over-fitted) curve by a decreased distance and hence give an incorrect small error.
This effect is illustrated in Fig.\,\ref{fig:crossval_illu} using 10 training and 200 test samples generated from a quadratic function plus Gaussian noise of standard deviation $\sigma=0.4$.
The mean square error (MSE) is given by the mean of the squared distances 
\mbox{$E=\parallel\hat{\vec{x}}-\vec{x}\parallel^2$} 
between the data points $\vec{x}$ and their projections $\hat{\vec{x}}$ onto the curve.
The over-fitted and the well-fitted or ideal model are compared by using the same test data set.
It turns out that the test error of the true original model (Fig.\,\ref{fig:crossval_illu}C) is almost three times larger than the test error of the overly complex model (Fig.\,\ref{fig:crossval_illu}B), which over-fits the data. 
Test set validation clearly favours the over-fitted model over the correct model, and hence fails to validate nonlinear PCA.
\\
To understand this contradiction, we have to distinguish between an 
error in supervised learning and the fulfilment of specific criteria 
in unsupervised learning. 
Test set validation works well for supervised methods because we measure the error as the difference from a known target 
(e.g.,~class labels).
Since in unsupervised methods the target 
(e.g.,~the correct component) is unknown, we optimize a specific criterion. 
In nonlinear PCA the criterion is to project the data by the shortest distance onto a curve. 
But a more complex over-fitted curve covers more data space and hence can also achieve a smaller error on test data than the true original curve.

\section{The nonlinear PCA model}
\label{sec:2}
\begin{figure}[t]
  \centerline{\includegraphics[width=0.7\linewidth]{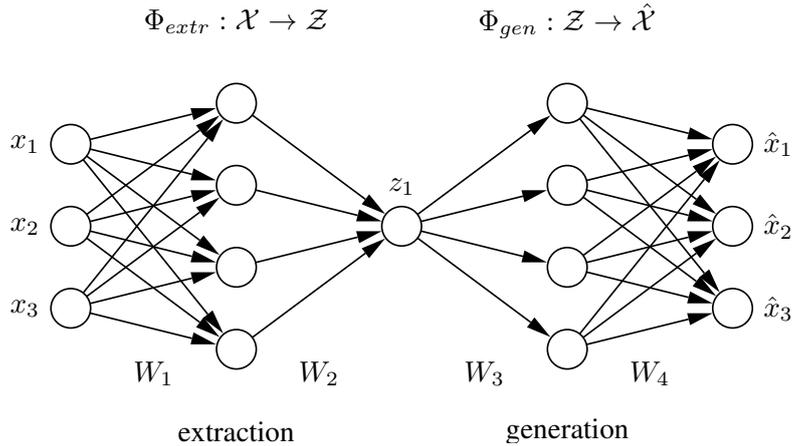}}
   \caption[abc]{
    The standard auto-associative neural network for nonlinear PCA.
    The network output $\hat{\vec{x}}$ is required 
    to approximate the input $\vec{x}$.
    Illustrated is a \mbox{3-4-1-4-3} network architecture.
    Three-dimensional samples $\vec{x}$ are compressed to one
    component value $\vec{z}$ in the middle by the extraction part.
    The inverse generation part
    reconstructs $\hat{\vec{x}}$ from $\vec{z}$. 
    The sample $\hat{\vec{x}}$ is a
    noise-reduced representation of $\vec{x}$, located on the component curve.
    }\label{fig:net_sym}
 \end{figure}

Nonlinear PCA (NLPCA) can be performed by using a multi-layer perceptron (MLP) of an auto-associative topology, also known as auto-encoder, replicator network, bottleneck, or sand-glass type network, see Fig.\,\ref{fig:net_sym}.
\\
The auto-associative network performs an identity mapping.
The output $\hat{\vec{x}}$ is forced to approximate the input $\vec{x}$ by minimising the squared reconstruction error
\mbox{$E=\parallel\hat{\vec{x}}-\vec{x}\parallel^2$}.
The network can be considered as consisting of two parts:
the first part represents the extraction function
\mbox{$\Phi_{extr} : {\mathcal X} \rightarrow {\mathcal Z}$},
whereas the second part represents the inverse function, the generation
or reconstruction function
\mbox{$\Phi_{gen} : {\mathcal Z} \rightarrow \hat{\mathcal X}$}.
A hidden layer in each part enables the network to perform nonlinear
mapping functions.
By using additional units in the component layer in the middle, 
the network can be extended to extract more than one component. 
Ordered components can be achieved by using a hierarchical nonlinear PCA\,\cite{SchVig02}.
\\
For the proposed validation approach, we have to adapt nonlinear PCA to be able to estimate missing data.
This can be done by using an inverse nonlinear PCA model\,\cite{SchKapGuy05} which optimises the generation function by using only the second part of the auto-associative neural network. 
Since the extraction mapping ${\mathcal X} \rightarrow {\mathcal Z}$ is lost, we have to estimate both the weights $\vec{w}$ and also the inputs $\vec{z}$ which represent the values of the nonlinear component. Both $\vec{w}$ and $\vec{z}$ can optimised simultaneously to minimise the reconstruction error, as shown in\,\cite{SchKapGuy05}. 
\\
The complexity of a model can be controlled by a weight-decay penalty term\,\cite{Hin87}
added to the error function 
$E_{total}=E+\nu \left( \sum_iw_i^2 \right)$, $w$ are the network weights. By varying the coefficient $\nu$, the impact of the weight-decay term can be changed and hence we modify the complexity of the model which defines the flexibility of the component curves in nonlinear PCA.

\section{The missing data validation approach}
Since classical test set validation fails to select the optimal nonlinear PCA model, as illustrated in Fig.\,\ref{fig:crossval_illu}, I propose to evaluate the complexity of a model by using the error in missing 
data estimation as the criterion for model selection.
This requires to adapt nonlinear PCA for missing data as done in the inverse nonlinear PCA model\,\cite{SchKapGuy05}.
The following model selection procedure can be used to find the optimal weight-decay complexity parameter $\nu$ of the nonlinear PCA model:\\
1.~Choose a specific complexity parameter $\nu$.\\  
2.~Apply inverse nonlinear PCA to a training data set.\\
3.~Validate the nonlinear PCA model by its performance on missing data estimation of 
an independent test set in which one or more elements $x_i^n$ of a sample $\vec{x}^n$ are randomly rejected.
The mean of the squared errors 
\mbox{$e_i^n=\parallel\hat{x}_i^n-x_i^n\parallel^2$}
between the randomly removed values $x_i^n$ and their estimations $\hat{x}_i^n$ by the nonlinear PCA model is used as the validation or generalization error.\\
Applied to a range of different weight-decay complexity parameters $\nu$, the optimal model complexity~$\nu$ is given by the lowest missing value estimation error. 
To get a more robust result, for each complexity setting, nonlinear PCA can be repeatedly applied by using different weight-initializations of the neural network. The median can then be used for validation as shown in the following examples.

\section{Validation examples}
\label{sec:3}
The first example of a nonlinear data set shows that model validation based on missing data estimation performance provides a clear optimum of the complexity parameter. The second example demonstrates that the proposed validation ensures that nonlinear PCA does not describe data in a nonlinear way when the inherent data structure is, in fact, linear.

\subsection{Helix data}

\begin{figure*}[t]
  \centerline{\includegraphics[width=0.7\linewidth]{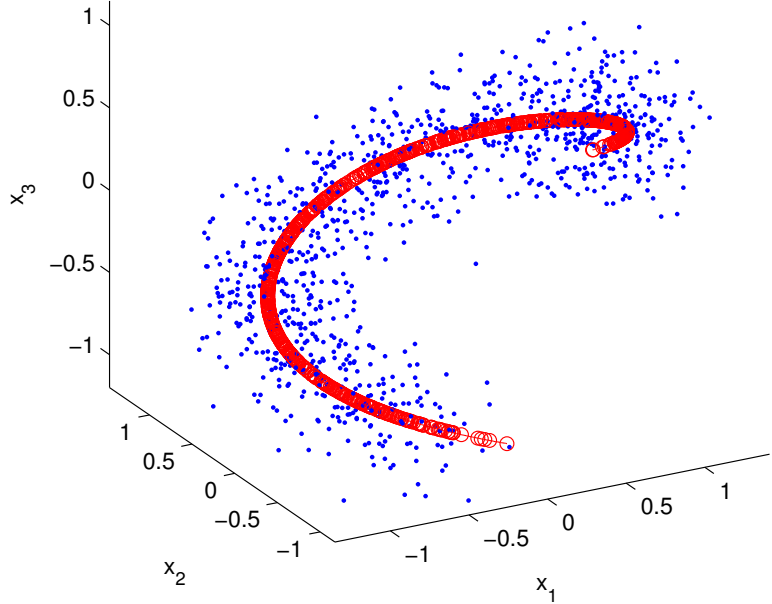}}
   \caption[abc]{
    {\bf Helical data set.}
    Data are generated from a one dimensional helical loop 
    embedded in three dimensions and additive Gaussian noise. 
    The nonlinear component is plotted as a red line. Red 
    circles show the projections of the data (blue dots) onto the curve. 
    }\label{fig:helix}
 \end{figure*}

\begin{figure}[t]
  \centerline{\includegraphics[width=0.72\linewidth]{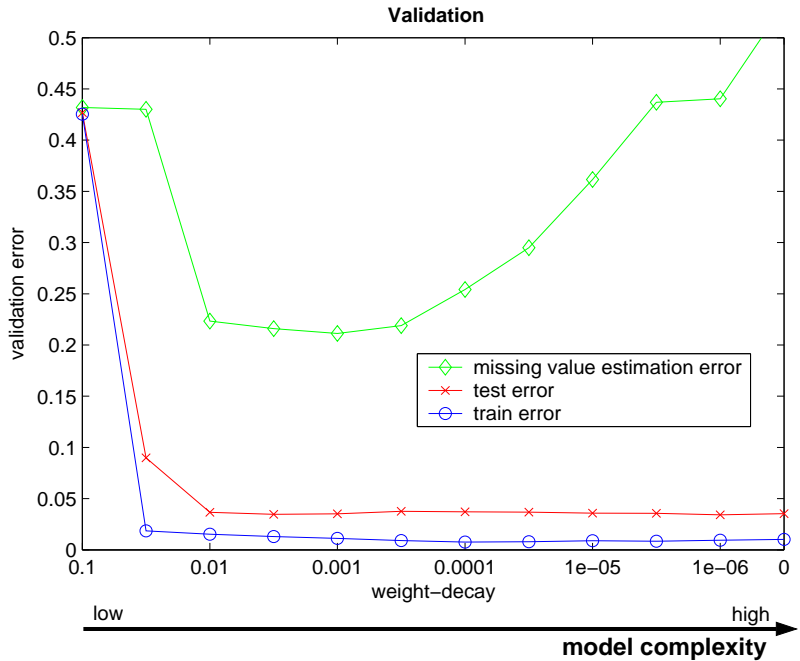}}
   \caption[abc]{
    {\sffamily\bfseries Model selection.}
    Missing data estimation is compared 
    with standard test set validation by using the helical 
    data set (Fig.\,\ref{fig:helix}).
    A nonlinear PCA network model of low complexity which is 
    almost linear (left) results 
    in a high error as expected for both the training and the test data.
    Only the missing data approach shows the 
    expected increase in validation error for over-fitted models (right). 
    }\label{fig:crossval_errorseries}
 \end{figure}

The nonlinear data set consist of data $\vec{x}=(x_1,x_2,x_3)^T$ that lie on a one-dimensional manifold, 
a helical loop, embedded in three dimensions, plus Gaussian noise~$\eta$ 
of standard deviation $\sigma=0.4$\,, 
as illustrated in Fig.\,\ref{fig:helix}. The samples $\vec{x}$ were generated 
from a uniformly distributed factor $t$ over the range [-0.8,0.8], $t$ represents 
the angle: 
  \[\begin{array}[t]{lclll}
  x_1 & = & \sin(\pi t) & + & \eta \\
  x_2 & = & \cos(\pi t) & + & \eta \\
  x_3 & = & t           & + & \eta
  \end{array}
  \]
Nonlinear PCA is applied by using a \mbox{1-10-3} network architecture optimized in 5,000 iterations by using the \emph{conjugate gradient descent} algorithm\,\cite{HesSti52}. 
\\
To evaluate different weight-decay complexity parameters $\nu$, nonlinear PCA is applied to 20 complete samples 
generated from the helical loop function and validated by using a missing data set of 1,000 incomplete 
samples in which randomly one value of the three dimensions is rejected per sample and can be easily estimated from the other two dimensions when the nonlinear component has the correct helical curve.
For comparison with standard test set validation, the same 1,000 (complete) samples are used.
This is repeatedly done 100 times for each model complexity with newly generated data each time. The median of missing data estimation over all 100 runs is finally taken to validate a specific model complexity.
\\
Fig.\,\ref{fig:crossval_errorseries} shows the results of comparing the proposed model selection approach with standard test set validation. 
It turns out that only the missing data approach is able to show a clear minimum in the performance curve. Test set validation, by contrast, shows a small error even for very complex (over-fitted) models.
This is contrary to our experience with supervised learning, 
where the test error becomes large again when the model over-fits.
Thus, test set validation cannot be used to determine the optimal model complexity of unsupervised methods.
In contrast, the missing value validation approach shows that the optimal complexity setting of the weight-decay coefficient is in the range $0.01 < \nu < 0.0001$.

\subsection{Linear data}
\begin{figure}[t]
  \centerline{\includegraphics[width=0.85\linewidth]{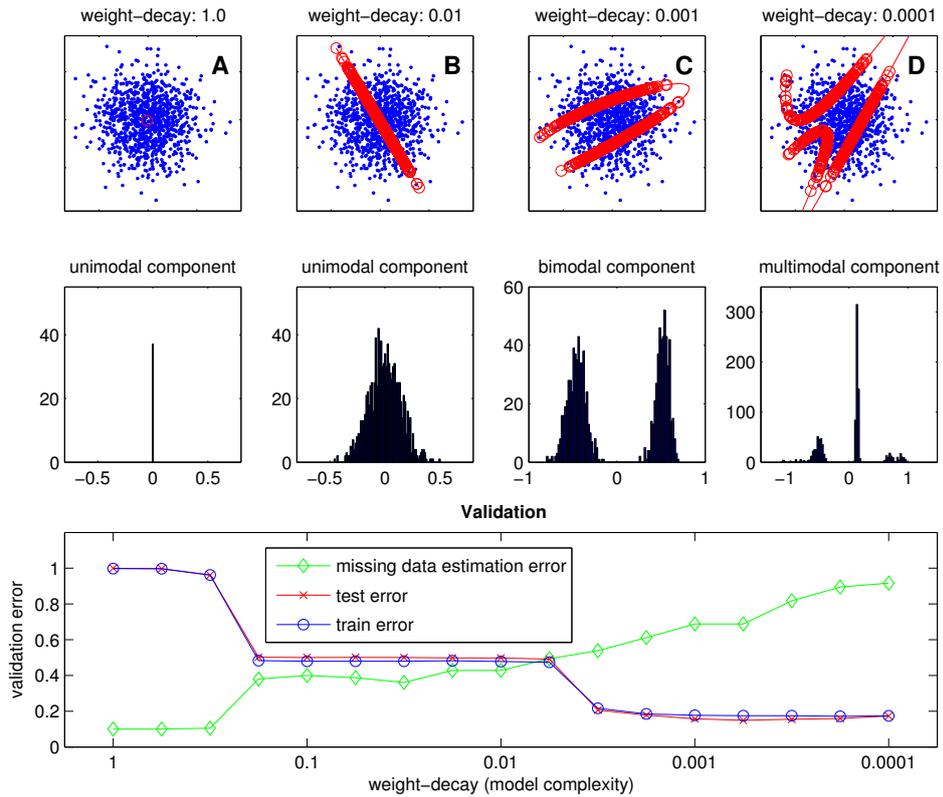}}
   \caption[abc]{
    Nonlinear PCA is applied to data of a two-dimensional 
    Gaussian distribution.
    ({\sffamily\bfseries C,D}) Over-fitted models.
    ({\sffamily\bfseries B}) A weight-decay of 0.01 forces nonlinear PCA to 
    describe the data by a linear component.
    ({\sffamily\bfseries A}) An even stronger penalty of 1.0 forces 
    a single point solution. 
    Below: Only the missing data approach shows that it would be 
    best to impose a very strong penalty that forces the network 
    into a linear solution.
    }\label{fig:GaussData}
 \end{figure}
 
Nonlinear PCA can also be used to answer the question of whether high-dimensional observations are driven by an unimodal or a multimodal process, e.g., in atmospheric science for analysing the El Ni\~no-Southern Oscillation\,\cite{Hsi01}.
But applying nonlinear PCA can be misleading if the model complexity is insufficiently controlled: multimodality can be incorrectly detected in data that are inherently unimodal, as pointed out by Christiansen\,\cite{Chr05}.
Fig.\,\ref{fig:GaussData}\,C\,\&\,D illustrates that if the model complexity is too high, even linear data is described by nonlinear components.
Therefore, to obtain the right description of the data, controlling the model complexity is very important.
Fig.\,\ref{fig:GaussData} shows the validation error curves of the standard test set and the proposed missing data validation for different model complexities. The median of 500 differently initialized 1-4-2 networks is plotted. Again, it is shown that standard test set validation fails in validating nonlinear PCA. 
With increasing model complexity, classical test set validation shows an decreasing error, and hence favours over-fitted models.  
By contrast, the missing value estimation error shows correctly that the optimum would be a strong penalty which gives a linear or even a point solution, thereby confirming the absence of nonlinearity in the data.
This is correct because the data consists, in principle, of Gaussian noise centred at 
the point~(0,0).\\
While test set validation favours over-fitted models which produce components that incorrectly show multimodal distributions, missing data validation confirms the unimodal characteristics of the data. Nonlinear PCA in combination with missing data validation can therefore be used to find out whether a high-dimensional data set is generated by a unimodal or a multimodal process.

\section{Test set versus missing data approach}
\label{sec:4} 
In standard test set validation, the nonlinear PCA model is trained using a training set ${\mathcal X}$. An independent test set ${\mathcal Y}$ is then used to compute a validation error as 
\mbox{$E=\parallel\hat{\vec{y}}-\vec{y}\parallel^2$}, where $\hat{\vec{y}}$ is the output of the nonlinear PCA given the test data $\vec{y}$ as the input. The test set validation reconstructs the test data from the test data itself. The problem with this approach is that increasingly complex functions can give approximately $\hat{\vec{y}}=\vec{y}$, thus favouring complex models.
While test set validation is a standard approach in supervised applications, 
in unsupervised techniques it suffers from the lack of a known target (e.g., a class label). 
Highly complex nonlinear PCA models, which over-fit the original training data, 
are in principle also able to fit test data better 
than would be possible by the true original model.
With higher complexity, a model is able to describe a more
complicated structure in the data space.
Even for new test samples, it is more likely to find a short projecting
distance (error) onto a curve which covers the 
data space almost complete
than by a curve of moderate complexity
(Fig.\,\ref{fig:crossval_illu}).
The problem is that we can project the data onto \emph{any}
position on the curve. There is no further restriction in pure test set
validation.
In missing data estimation, by contrast, the required position on the
curve is \emph{fixed}, given by the remaining available
values of the same sample.
The artificially removed missing value of a test sample gives an exact target 
which have to be predicted from the available values of the same test sample.
While test set validation predicts the test data from the test data itself, the missing data validation predicts removed values from the remaining values of the same sample.
Thus, we transform the unsupervised validation
problem into a kind of supervised validation problem.

\section{Conclusion}
\label{sec:5}
In this paper, 
the missing data validation approach to model selection is proposed to be applied to 
the auto-associative neural network based nonlinear PCA. The idea behind this approach is that the true generalization error in unsupervised methods is given by a missing value estimation error and not by the classical test set error.
The proposed missing value validation approach can therefore be seen as an adaptation of the standard test set validation so as to be applicable to unsupervised methods.
The absence of a target value in unsupervised methods is replaced by using artificially removed missing values as expected target values that have to be predicted from the remaining values of the same sample. It can be shown that standard test set validation clearly fails to validate nonlinear PCA. In contrast, the proposed missing data validation approach was able to validate correctly the model complexity.

\section*{Availability of Software}
\label{sec:6}
  A MATLAB\textsuperscript{\textregistered} implementation of nonlinear PCA including the inverse model for estimating missing data is available at:\\
\href{http://www.NLPCA.org/matlab.html}{\texttt{http://www.NLPCA.org/matlab.html}}
\\
An example of how to apply the proposed validation approach 
can be found at:\linebreak
\href{http://www.NLPCA.org/validation.html}{\texttt{http://www.NLPCA.org/validation.html}}

\bibliographystyle{unsrtnat}  
\bibliography{scholz}

\end{document}